\pdfoutput=1

\documentclass[11pt]{article}

\usepackage{ACL2023}

\usepackage{times}
\usepackage{latexsym}
\usepackage{tabularx}
\usepackage{booktabs}
\usepackage[T1]{fontenc}
\usepackage[utf8]{inputenc}
\usepackage{amsmath, amssymb}
\usepackage{microtype}
\usepackage{tabularx}
\usepackage{hyperref}
\usepackage{inconsolata}
\usepackage{graphicx}
\usepackage{stfloats}
\usepackage{multirow}
\newcolumntype{Y}{>{\centering\arraybackslash}X}
\usepackage{enumitem}

\title{Exploring Selective Retrieval-Augmentation for Long-Tail Legal Text Classification}

\author{
  Boheng Mao \\
  University of Cambridge \\
  Cambridge, United Kingdom \\
  \texttt{bm687@cam.ac.uk} \\
}

\begin{document}
\maketitle
\begin{abstract}
Legal text classification is a fundamental NLP task in the legal domain. Benchmark datasets in this area often exhibit a long-tail label distribution, where many labels are underrepresented, leading to poor model performance on rare classes. This paper explores Selective Retrieval-Augmentation (SRA) as a proof-of-concept approach to this problem. SRA focuses on augmenting samples belonging to low-frequency labels in the training set, preventing the introduction of noise for well-represented classes, and requires no changes to the model architecture. Retrieval is performed only from the training data to ensure there is no potential information leakage, removing the need for external corpora simultaneously. SRA is tested on two legal text classification benchmark datasets with long-tail distributions: LEDGAR (single-label) and UNFAIR-ToS (multi-label). Results show that SRA achieves consistent gains in both micro-F1 and macro-F1 over LexGLUE baselines. The code repository is available at: \url{https://github.com/Boheng-Mao/sra-legal}
\end{abstract}

\section{Introduction}
As a core task in legal natural language processing (NLP), legal text classification provides support to various downstream applications \cite{chalkidis-etal-2022-lexglue}. Examples include contract analysis and the retrieval of legal documents \cite{hendrycks2021cuad}, which depend on precise classification of pertinent texts. However, many benchmark legal datasets such as LEDGAR and UNFAIR-ToS contain skewed, long-tail label distributions \citep{lippi2019claudette, tuggener-etal-2020-ledgar}, suggesting that some categories in such datasets appear more commonly than others. Due to such an imbalance, models often perform well on common categories but struggle on rare ones, affecting the overall performance \citep{he2009learning, johnson2019survey}.

\begin{figure}[t]
    \centering
    \includegraphics[width=\columnwidth]{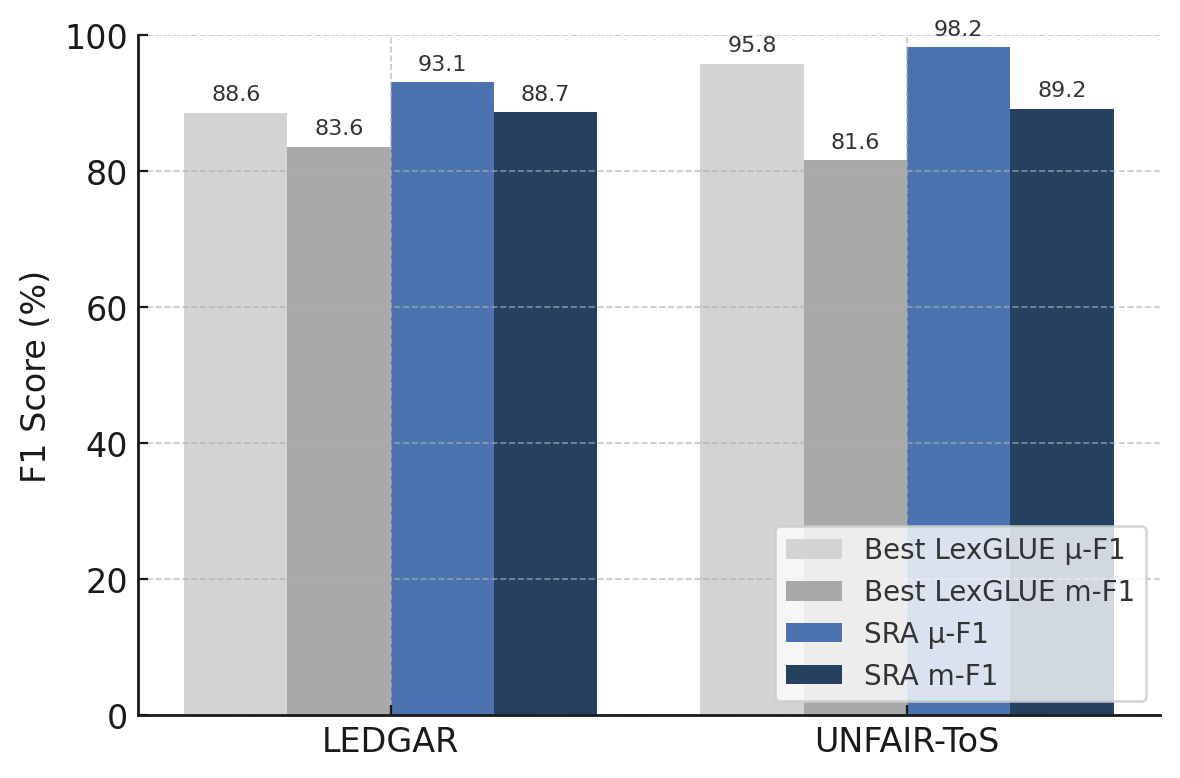}
    \caption{Overall performance comparison between SRA and the best-performing LexGLUE baselines on LEDGAR and UNFAIR-ToS. Results show consistent improvements in both micro-F1 ($\mu$-F1) and macro-F1 ($m$-F1)}
    \label{fig:overall}
\end{figure}

Retrieval-augmented techniques have been applied to text classification tasks \cite{lewis2020retrieval, yu2023retrieval}. For instance, \citet{abdullahi2024retrieval} introduced a retrieval-augmented zero-shot system that enhances input queries with external knowledge sources such as Wikipedia. \citet{chalkidis2023retrieval} demonstrated that in semantically rich fields, like law and biomedicine, combining retrieved information with the input strengthens the performance of multi-label text classification models. However, these approaches typically apply augmentation to all samples, and some depend on external corpora \cite{karpukhin-etal-2020-dense}. This motivates the creation of more straightforward techniques that can address these limitations while concentrating exclusively on the legal field.

This paper explores Selective Retrieval-Augmentation (SRA) for long-tail legal text classification. SRA focuses on underrepresented labels and retrieves semantically relevant content entirely from the training set. Due to their length, semantic richness, and self-contained nature, legal documents are particularly well-suited for in-domain retrieval without relying on external knowledge \citep{ariai2024natural}. Augmentation is applied consistently during training, validation, and testing, while the classifier architecture remains unchanged. 
The evaluation setup deliberately assumes access to ground-truth labels to isolate the effect of selectively augmenting low-frequency classes. The limitations of this design are discussed later in the paper.

The contributions of this paper are:
\paragraph{Methodological:} Explore Selective Retrieval-Augmentation (SRA), a strategy that targets low-frequency classes for augmentation to reduce class imbalance in supervised legal text classification.
\paragraph{Domain-specific:} Demonstrate that legal texts, which are comprehensive and semantically rich, are suitable for in-domain retrieval, which allows for augmentation without the need for an external knowledge source.
\paragraph{Empirical:} Implement SRA with a RoBERTa-base model \cite{liu2019roberta} and evaluate it on two legal benchmarks: LEDGAR (single-label) and UNFAIR-ToS (multi-label). As illustrated in Figure~\ref{fig:overall}, SRA achieves consistent gains over LexGLUE baselines across both datasets \cite{chalkidis-etal-2022-lexglue}.

\section{Related Work}

\subsection{Legal Benchmarks and Long-tail Classification}
A significant challenge in legal text classification is the long-tail distribution of label frequencies, exemplified by benchmarks such as LEDGAR, UNFAIR-ToS, and EUR-Lex \cite{chalkidis2019large}. In the vast machine learning literature, methodologies such as logit adjustment \citep{menon2021logit} and decoupled representation learning \citep{kang2020decoupling} have been introduced, with recent surveys providing comprehensive analyses of long-tailed classification strategies \citep{de2024survey}. Although these techniques alleviate imbalance at the loss or classifier level, they do not provide direct contextual support for infrequent classes. In contrast, Selective Retrieval-Augmentation (SRA) explores the idea of selectively incorporating in-domain examples for underrepresented labels, offering a complementary perspective to existing approaches.

\subsection{Retrieval-Augmented Classification}
Retrieval-augmented approaches were initially proposed as supplements to large language models in knowledge-intensive tasks by retrieving suitable information from external sources \cite{lewis2020retrieval}. The concept has since been extended to text classification, including frameworks developed for multi-label settings \cite{chalkidis2023retrieval}, as well as approaches targeting few-shot \cite{yu2023retrieval} and zero-shot classification \cite{abdullahi2024retrieval}. These studies have shown the flexibility of retrieval augmentation, but they rarely focus on underrepresented labels in skewed datasets, and many are designed for open-domain scenarios where external information is essential. Building on this line of work, Selective Retrieval-Augmentation (SRA) explores the use of in-domain retrieval within the training set to support less common categories, illustrating a way to provide augmentation without relying on outside sources.

\begin{figure}[b]
    \centering
    \includegraphics[width=\columnwidth]{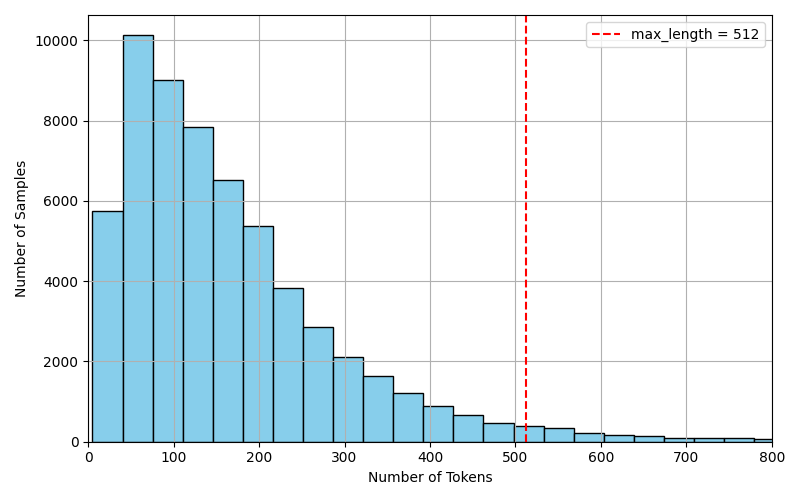}
    \caption{Token length distribution of texts in LEDGAR's augmented train set (65\% cutoff). The majority of samples have lengths below the maximum of 512 tokens.}
    \label{fig:token_length_distribution}
\end{figure}

\section{Methodology}
\label{sec:3}

\subsection{Selective Retrieval}
The goal of SRA is to detect low-frequency classes in the training set and return semantically associated examples only for instances belonging to such classes.
Formally, let $\mathcal{D} = \{(x_i, y_i)\}_{i=1}^N$ denote the training set with label set $\mathcal{C}$, and let $f(c)$ be the frequency of class $c \in \mathcal{C}$. 
The set of low-frequency classes $\mathcal{C}_{\text{low}}$ is obtained by ranking all labels in ascending order of $f(c)$ and selecting the bottom $\alpha$ proportion, where $\alpha \in (0,1]$ is a predefined cutoff:
\[
\mathcal{C}_{\text{low}} = \{ c_{(1)}, \ldots, c_{(k)} \},\quad
k = \lfloor \alpha \, |\mathcal{C}| \rfloor.
\]

For each sample $x$ with $y \in \mathcal{C}_{\text{low}}$, retrieval is performed in two stages: (1) TF-IDF ranking for efficient candidate selection \cite{singhal2001modern}, and (2) semantic re-ranking with Sentence-BERT embeddings \cite{reimers2019sentence} using cosine similarity. 
The top-$k$ candidates (default $k{=}1$) are collected, excluding any self-retrieval.

Lastly, the retrieved examples $\{r_1, \dots, r_k\}$ are concatenated with the original text using a short prompt:
$p_{\text{orig}}=\texttt{``Original clause:''}$ and 
$p_{\text{ref}}=\texttt{``Related clause for reference:''}$. 
The augmented input is
\[
x' = p_{\text{orig}} \oplus x \oplus \texttt{[SEP]} \oplus p_{\text{ref}} \oplus r_1 \oplus \dots \oplus r_k .
\]
so that the model interprets them as additional evidence rather than contents in the original input. 

\subsection{Augmentation Strategy}
The definition of $\mathcal{C}_{\text{low}}$ is derived solely from the training set distribution. 
Once determined, this set of uncommon labels is fixed and applied consistently across training, validation, and test splits. 
For all splits, retrieval is performed only from the training set. This ensures augmentation always returns in-domain examples from seen data (training set) and prevents information leakage. 

To control the length of the input, each retrieved clause is truncated to at most 64 tokens. The concatenated sequence is then limited to the maximum model length (512 tokens for RoBERTa-base). Figure~\ref{fig:token_length_distribution} shows that for LEDGAR, 95\% of augmented samples contain fewer than 421 tokens and 99\% have fewer than 643 tokens. In UNFAIR-ToS, texts are much shorter, so truncation rarely occurs. Thus, truncation only affects a small number of samples, showing that most of the important information is kept.

The SRA framework is a data-level augmentation strategy and does not change the original model architecture. In this work, a RoBERTa-base model is used as the classification backbone. The model uses the standard RoBERTa encoder to encode the augmented sequence $x'$ based on the input $x$. Finally, the output is passed through a linear classification layer:
\[
h = \text{RoBERTa}(x'), \quad 
\hat{y} = \text{softmax}(Wh + b).
\]
This design ensures full compatibility with the usual training procedures. The classifier learns to utilize retrieved clauses as extra information for low-frequency labels during training.

\begin{table*}[t]
\centering
\footnotesize
\renewcommand{\arraystretch}{1.15}
\begin{tabularx}{\textwidth}{l l Y Y}
\toprule
\textbf{Dataset} & \textbf{Method} & \textbf{Micro-F1} & \textbf{Macro-F1} \\
\midrule
\multirow{4}{*}{LEDGAR}
  & Baseline (RoBERTa-base)   & 0.879 & 0.827 \\
  & Full Retrieval-Augmentation & 0.875 & 0.816 \\
  & \textbf{SRA} (65\%)       & \textbf{0.931} & \textbf{0.887} \\
  & Best LexGLUE baseline (RoBERTa-large) & 0.886 & 0.836 \\
\midrule
\multirow{5}{*}{Unfair-ToS}
  & Baseline (RoBERTa-base)   & 0.952 & 0.807 \\
  & Full Retrieval-Augmentation & 0.953 & 0.790 \\
  & \textbf{SRA} (Non-empty labels only) & \textbf{0.988} & \textbf{0.924} \\
  & SRA (Low-frequency \& non-empty labels only) & 0.968 & 0.889 \\
  & Best LexGLUE baseline (Legal-BERT) & 0.960 & 0.830 \\
\bottomrule
\end{tabularx}
\caption{
Test performance under the best augmentation cutoff per dataset (selected on the validation set; see Sec~\ref{sec:4.5} for the ablation). 
Numbers for “Best LexGLUE baseline” are taken from the official LexGLUE leaderboard. “Low-frequency” in Unfair-ToS refers to labels with frequency $\leq 76$ in the training set.
}
\label{tab:main-results}
\end{table*}

\section{Experiments}
\subsection{Datasets}
Experiments are conducted on datasets from the LexGLUE benchmark, which is a widely used set of datasets for legal NLP model evaluation \cite{chalkidis-etal-2022-lexglue}. LEDGAR \cite{tuggener-etal-2020-ledgar} and UNFAIR-ToS \cite{lippi2019claudette} are chosen due to their long-tail distributions and applicability to both single-label and multi-label contexts, rendering them ideal for SRA evaluation. Table~\ref{tab:datasets} provides a summary of the dataset statistics.

LEDGAR is constructed with contract clauses that have been appropriately categorized with one of 100 possible functional categories (single-label classification). UNFAIR-ToS includes Terms of Service (ToS) clauses that have been designated with up to eight distinct forms of unfair contract practices (multi-label classification). A significant portion of ToS clauses do not include any unfairness, resulting in the assignment of empty label sets, effectively introducing an additional class.

The two datasets illustrate different long-tail patterns. In LEDGAR, imbalance is caused by class frequencies that are extremely dispersed, with many labels appearing only a handful of times (see Appendix~\ref{app:dataset} for a full distribution). On the other hand, UNFAIR-ToS has a different type of imbalance: the dataset is dominated by empty labels. 88.6\% of the training set labels, 89.9\% of the validation set labels, and 89.3\% of the test set labels are empty. Positive cases are rare, causing a long-tail distribution of non-empty labels. 

Subsequently, different SRA strategies are applied to each dataset based on its distributional characteristics. 
On LEDGAR, augmentation is aimed at low-frequency classes. On UNFAIR-ToS, it naturally coincides with enhancing all non-empty cases since empty-label samples compose the majority of the distribution.

\begin{table}[t]
\centering
\footnotesize
\setlength{\tabcolsep}{4pt}
\renewcommand{\arraystretch}{1.05}
\begin{tabularx}{\columnwidth}{lccc}
\toprule
\textbf{Dataset} & \textbf{Task} & \textbf{Classes} & \textbf{Train / Val / Test} \\
\midrule
LEDGAR     & Single-label  & 100 & $\sim$60k / 10k / 10k \\
UNFAIR-ToS & Multi-label  & 8 + 1 & $\sim$5.5k / 2.3k / 1.6k \\
\bottomrule
\end{tabularx}
\caption{
Datasets from the LexGLUE benchmark used in experiments.  
UNFAIR-ToS defines 8 unfairness categories, but the majority of samples have no label (empty).
}
\label{tab:datasets}
\end{table}

\subsection{Implementation Details}
The model used for SRA experiments on both datasets is RoBERTa-base with a linear classification head. The model is full-finetuned using the AdamW optimizer \cite{loshchilov2017decoupled} with an initial learning rate of 3e-5, a linear decay scheduler with no warmup and no weight decay, and a maximum gradient norm of 1.0. The batch size is set to 32 for training and 64 for evaluation. Early stopping with patience = 3 is applied based on the validation micro-F1 score.

The training setup follows the official LexGLUE baseline configuration, with the only modification being the batch size (32 for training and 64 for evaluation, instead of 8 for both) to accelerate training. This change does not affect performance, verified by re-running the RoBERTa-base baseline with a batch size of 8, which showed nearly identical results compared to the outcome of RoBERTa-base in LexGLUE.

Retrieval follows the procedure described in Section~\ref{sec:3}, and is applied to training, validation, and test sets for consistency across splits (see Appendix~\ref{app:setting} for detailed configuration).

Performance is evaluated using micro-F1 and macro-F1. Both micro-F1 and macro-F1 are reported: micro-F1 reflects the overall performance across all classes, while macro-F1 captures improvements on uncommon classes. Macro-F1 is considered the primary metric, as it better reflects the effectiveness of selective augmentation under long-tail distributions.

\subsection{Baselines}
Three configurations are considered in the experiments:
\begin{enumerate}
    \item Baseline: Full-finetune the RoBERTa-base without retrieval augmentation.
    \item Full Retrieval-Augmentation: Retrieval augmentation applied to all samples, regardless of class frequency.
    \item Selective Retrieval-Augmentation (SRA): The proposed method, where only samples corresponding to low-frequency classes in the training set are augmented.
\end{enumerate}

In addition, the performances of the best baseline models reported in the LexGLUE benchmark are included to ensure comparability with prior work.

\subsection{Main Results}
The augmentation ratio, i.e., the proportion of low-frequency classes $\mathcal{C}_{\text{low}}$ that are augmented, is treated as a hyperparameter. Different ratios are tested on the validation set, and the best-performing cutoff is applied to the test set for reporting results. Table~\ref{tab:main-results} summarizes the results (see Section~\ref{sec:4.5} for a full ablation). Several patterns can be observed:

\begin{figure*}[t]
    \centering
    \includegraphics[width=\textwidth]{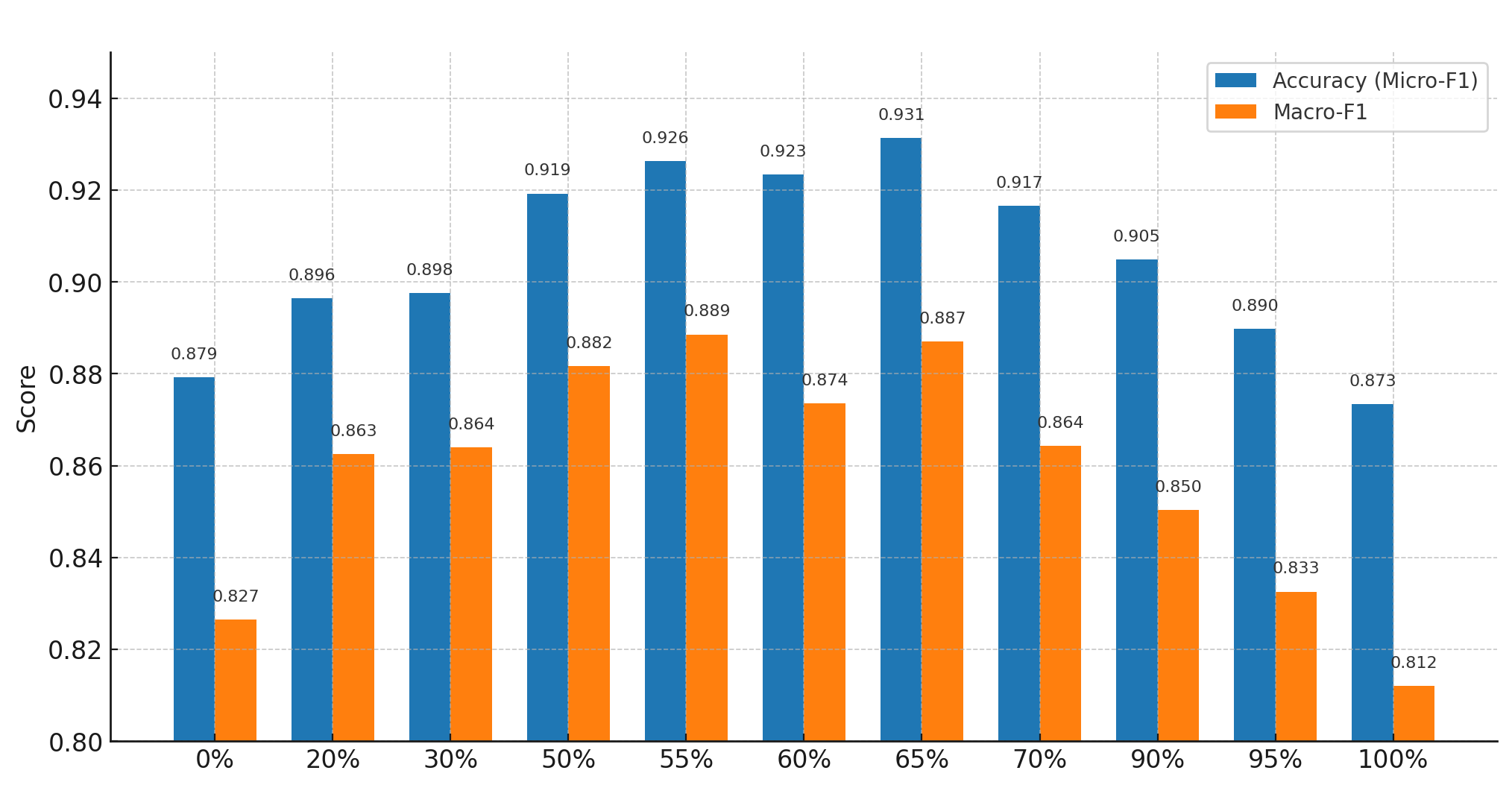}
    \caption{
    Ablation on LEDGAR: Accuracy (Micro-F1) and Macro-F1 vs. cutoff ratio. 
    Selective augmentation is most effective between 55\%–65\%. 
    }
    \label{fig:cutoff-ablation}
\end{figure*}

\begin{enumerate}[topsep=0pt,itemsep=0pt,parsep=0pt,partopsep=0pt]
    \item \textbf{LEDGAR (single-label)}. The baseline RoBERTa-base achieves 0.879 micro-F1 and 0.827 macro-F1. Full Retrieval-Augmentation degrades performance (macro-F1 0.816), indicating that indiscriminate augmentation injects noise. SRA with a 65\% cutoff delivers the best results (0.931 / 0.887), improving macro-F1 by +6.0\% over the baseline and surpassing the best LexGLUE baseline (RoBERTa-large).
    \item \textbf{UNFAIR-ToS (multi-label)}. The baseline RoBERTa-base reaches 0.952 micro-F1 and 0.807 macro-F1. Full Retrieval-Augmentation slightly lowers performance (macro-F1 0.790), while SRA applied to non-empty labels yields the best results (0.988 / 0.924), giving large gains over both the baseline and the best LexGLUE model (Legal-BERT).
\end{enumerate}
\noindent
These results indicate that selective augmentation, as opposed to indiscriminate retrieval, is crucial for effective legal text classification under long-tail distributions.

\subsection{Ablation Studies}
\label{sec:4.5}
To better understand the effect of the augmentation ratio, different cutoffs for $\mathcal{C}_{\text{low}}$ are tested. Figure~\ref{fig:cutoff-ablation} shows the test results on LEDGAR as the ratio varies from 0\% (baseline) to 100\% (full retrieval augmentation). 

On LEDGAR, performance improves steadily as the ratio increases, but drops once augmentation becomes too extensive. The maximum lies between 55\% and 65\%, where macro-F1 is maximized. This demonstrates that selective augmentation is more effective than augmenting all classes under LEDGAR's long-tailed distribution.  

On UNFAIR-ToS, the skewness comes from the large amount of empty-label samples. In this case, the best performance results from recognizing "selective" as "selecting all non-empty labels". However, augmenting all samples (full retrieval augmentation) lowers macro-F1, which is in accordance with what was observed on LEDGAR, where full augmentation introduced unwanted noise.

These results highlight that the optimal augmentation strategy depends on the underlying form of imbalance in the datasets.

\section{Analysis}

The following tests are conducted on LEDGAR with cutoff ratio = 65\% (see Appendix~\ref{app:stats} for tests on other ratios). 
\paragraph{(a) Statistical significance.}
SRA improves macro-F1 by +0.060 over the baseline (0.827 $\rightarrow$ 0.887). 
A bootstrap confidence interval confirms the gain is robust (95\% CI = [0.0459, 0.0662]), and McNemar’s test gives $p \ll 0.01$, eliminating chance as a potential explanation.

\paragraph{(b) Bucketed performance.}
Grouping labels by training frequency (low/medium/high), the largest relative gains are observed in the rare and medium-frequency buckets.
Macro-F1 improves from 0.759 $\rightarrow$ 0.808 (+0.049) for the rare bucket and 0.827 $\rightarrow$ 0.905 (+0.078) for the medium bucket, while frequent classes also see consistent gains (0.904 $\rightarrow$ 0.948). 
This shows SRA effectively corrects long-tail imbalance without sacrificing performance on common classes.

\paragraph{(c) Top-k categories.}
Inspection of label-wise results reveals that some of the lowest-frequency classes (e.g., label IDs 4, 5, 28, 50, 60) achieve noticeable improvements, with macro-F1 gains exceeding +0.20 in several cases. 
For instance, label~4 improves from 0.34 $\rightarrow$ 0.76, demonstrating that retrieval can turn previously underperforming classes into well-represented ones.

\begin{figure}[t]
    \centering
    \includegraphics[width=\columnwidth]{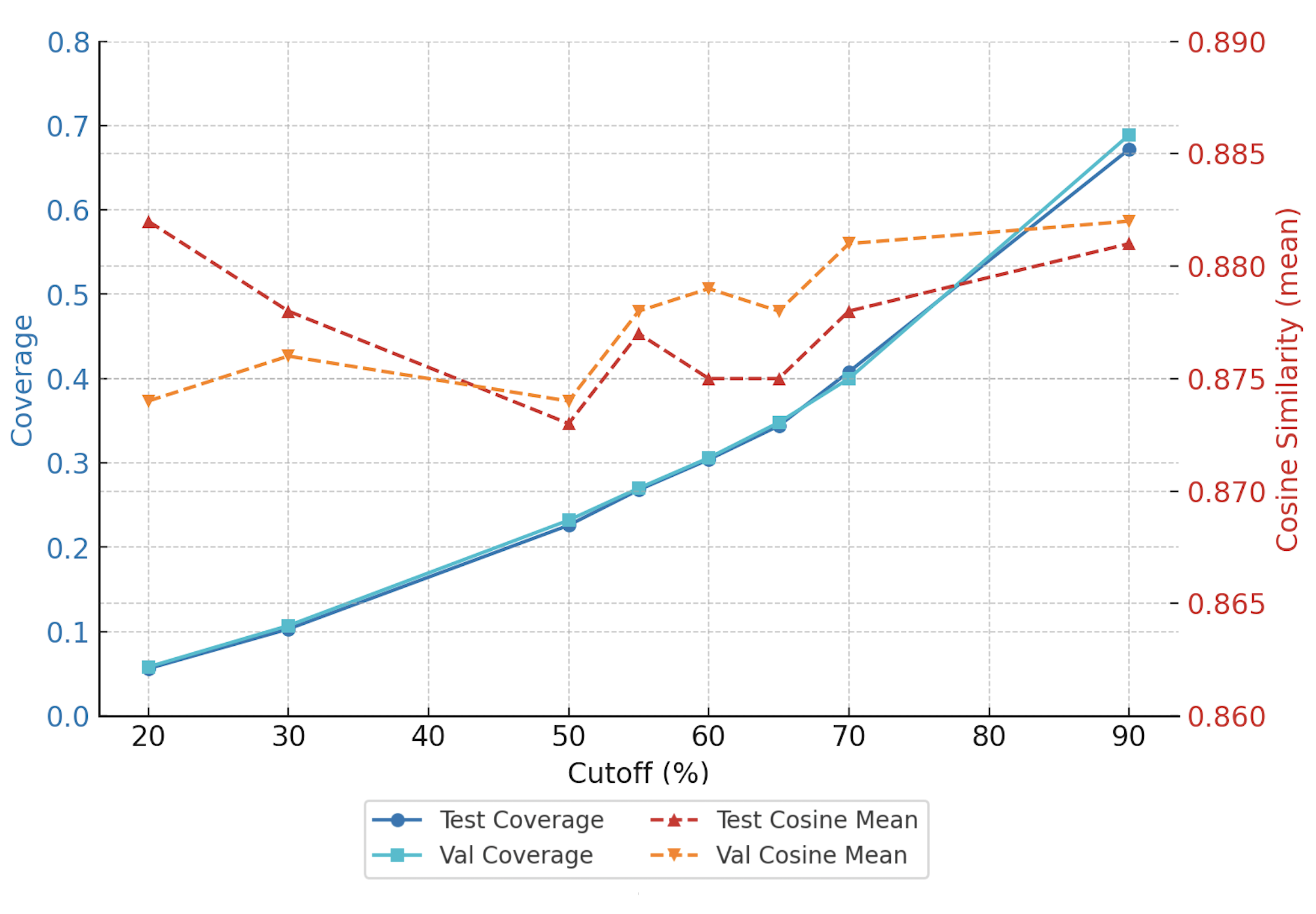}
    \caption{
    Retrieval coverage and cosine similarity under different augmentation cutoffs on LEDGAR.
    }
    \label{fig:coverage-similarity}
\end{figure}

\paragraph{(d) Coverage and similarity.}
Figure~\ref{fig:coverage-similarity} illustrates how different cutoffs for LEDGAR affect retrieval coverage and cosine similarity. As the cutoff increases, coverage improves steadily, yet similarity remains stable. This demonstrates that the primary reason for differences in performance is the number of samples that are augmented, rather than the quality of the retrieval itself.

For UNFAIR-ToS, applying SRA selectively to all non-empty instances gives the best results, reaching 0.988 micro-F1 and 0.924 macro-F1. This implies that although the imbalance appears in a different form, the same principle still holds: SRA benefits rare positive cases without harming the empty class.

\section{Conclusion}
This paper introduced Selective Retrieval-Augmentation (SRA) for long-tail legal text classification. The results show that directing augmentation toward tail classes can substantially improve both micro-F1 and macro-F1 compared with uniform augmentation and strong LexGLUE baselines. While the current implementation should be regarded as an exploratory study, the findings highlight the potential of tail-aware augmentation for addressing long-tail distributions in legal NLP.

\section*{Limitations}
The selective retrieval-augmentation strategy uses ground-truth labels during validation and testing to determine whether a sample belongs to the low-frequency set defined from the training distribution. This assumption is unrealistic for deployment, where labels are unavailable at inference time. In this work, it is adopted only to provide a proof-of-concept, showing that tail-aware augmentation can be beneficial. Future work may investigate label-free alternatives, for example uncertainty-based criteria or auxiliary models that estimate whether an input is likely to come from a tail class.

\section*{Ethics Statement}
This research utilizes publicly accessible legal text datasets (LEDGAR and UNFAIR-ToS) and does not incorporate personal or sensitive information. The suggested method aims to enhance classification accuracy for low-frequency categories, which may support fairer access to legal information. However, as with all machine learning approaches, potential biases in the datasets may be reflected in model outputs, and care should be taken before applying such models in real-world legal decision-making.

\bibliography{references}
\bibliographystyle{acl_natbib}

\appendix
\section*{Appendix}

\section{Dataset Statistics}
\label{app:dataset}
This appendix reports the label distributions of the two datasets used in this study. 
Both LEDGAR and UNFAIR-ToS exhibit highly imbalanced, long-tailed distributions, 
where a few labels dominate the data (see Table~\ref{tab:ledgar_buckets} and 
Table~\ref{tab:unfair_empty_nonempty}).

\begin{table*}[t]
\centering
\small
\begin{tabular*}{\textwidth}{@{\extracolsep{\fill}} l c c c c}
\toprule
Bucket & \#Classes & Avg.~Frequency & Min~Frequency & Max~Frequency \\
\midrule
High (Top 10\%)  & 10 & 1888.8 & 1166 & 3167 \\
Mid (Middle 40\%)& 40 &  682.1 &  434 & 1112 \\
Low (Bottom 50\%)& 50 &  276.5 &   23 &  419 \\
\bottomrule
\end{tabular*}
\caption{Bucketization of LEDGAR classes by frequency.}
\label{tab:ledgar_buckets}
\end{table*}

\begin{table*}[t]
\centering
\small
\begin{tabular*}{\textwidth}{@{\extracolsep{\fill}} l c c c}
\toprule
Split & Train & Validation & Test \\
\midrule
\% Samples with no unfairness label & 88.6\% & 89.9\% & 89.3\% \\
\% Samples with at least one unfairness label & 11.4\% & 10.1\% & 10.7\% \\
\bottomrule
\end{tabular*}
\caption{Distribution of empty vs.\ non-empty labels in the UNFAIR-ToS dataset. 
A large majority of samples contain no unfairness label, reflecting the long-tail property.}
\label{tab:unfair_empty_nonempty}
\end{table*}

\section{Retrieval Settings}
\label{app:setting}
This section summarizes the detailed configuration of the retrieval pipeline used in Selective Retrieval-Augmentation (SRA). 
All splits (train/validation/test) were processed with the same settings.

\paragraph{TF-IDF.} 
A scikit-learn \texttt{TfidfVectorizer} was used with word-level unigrams 
and bigrams. The minimum document frequency was set to 2, and documents 
appearing in more than 80\% of the corpus were discarded 
(\texttt{min\_df=2}, \texttt{max\_df=0.8}). Sublinear term frequency scaling 
was applied. For each query, the top-20 candidates were retrieved.

\paragraph{SBERT.} 
The dense encoder was \texttt{sentence-transformers/all-mpnet-base-v2}, downloaded from the Hugging Face Hub. Maximum sequence length was set to 256 tokens. Embeddings were computed in batches of 512 and normalized. For each query, the top-20 TF-IDF candidates were re-ranked with cosine similarity in the SBERT embedding space, and the top-5 were retained. From these, only the single best-retrieved clause (top-1) was used for augmentation. Each retrieved clause was truncated to 64 tokens.

\section{Additional Results on LEDGAR}
\label{app:stats}
Detailed results of SRA on the LEDGAR dataset under 
different cutoff values (20\%, 65\%, and 90\%) are included here.

\paragraph{20\% cutoff.} 
Macro-F1 gain: +0.0389 (95\% CI: [0.0285, 0.0493]), 
$p < 10^{-12}$. 
Bucket analysis shows the highest gains in low-frequency classes 
(+0.0915), while mid- and high-frequency classes improve by 
+0.0154 and +0.0070, respectively. 
Top-5 improved classes include ID 8, 5, 25, 60, 34. 

\paragraph{65\% cutoff.} 
Macro-F1 gain: +0.0566 (95\% CI: [0.0459, 0.0662]), 
$p < 10^{-82}$. 
Both mid- and high-frequency classes improve significantly (+0.078, +0.043), 
while low-frequency classes improve moderately (+0.049). 
Top-5 improved classes include ID 4, 60, 5, 50, 28. 

\paragraph{90\% cutoff.} 
Macro-F1 gain: +0.0211 (95\% CI: [0.0126, 0.0294]), 
$p < 10^{-21}$. 
All buckets improve slightly (+0.019, +0.022, +0.021). 
Top-5 improved classes include ID 4, 60, 52, 59, 7. 

\section{Case Study}
Although Selective Retrieval-Augmentation (SRA) improves performance overall, retrieval can sometimes introduce misleading signals. Two representative cases are provided here, with one failure from LEDGAR and one success from UNFAIR-ToS.

\paragraph{Case 1 (LEDGAR).}  
An instance where SRA misleads the model, but the baseline without SRA succeeds.  

\textbf{Original clause:}  
``Executive shall serve in an executive capacity and shall perform such duties as are customarily associated with his position, consistent with the bylaws or operating agreement of the Company and its Affiliates, as the case may be, and as reasonably required by the Board.''  

\textbf{Retrieved clause:}  
``During the Employment Term, the Executive shall serve as the President and Chief Executive Officer of the Company, reporting to the Company’s Board of Directors (the “Board”). In such position, the Executive shall have such duties, authority and responsibility as shall be determined from time to time by the Board.''  

\textbf{Outcome:} The baseline correctly predicted the gold label \textit{Duties} (Class~32).  
SRA instead predicted \textit{Positions} (Class~69), as the retrieved clause emphasized the executive’s formal title (``President and CEO''), introducing positional cues that overrode the duty-related context.  

\paragraph{Case 2 (UNFAIR-ToS).}  
An instance where SRA enables the correct prediction while the baseline without SRA fails.  

\textbf{Original clause:}  
``We reserve the right to remove content alleged to be infringing without prior notice, at our sole discretion, and without liability to you.''  

\textbf{Retrieved clause:}  
``We reserve the right to delete or disable content alleged to be infringing and terminate accounts of repeat infringers.''  

\textbf{Outcome:} The baseline predicted \emph{empty} (no label), whereas SRA correctly predicted \textit{Content removal} (Class~3).  
The retrieved clause reinforced removal cues (``remove/delete content'', ``infringing'', ``repeat infringers''), guiding the model toward the correct label.

\end{document}